\documentclass{bmvc2k}
\usepackage{pifont}
\usepackage{xcolor}
\usepackage{booktabs}
\usepackage{multirow}
\usepackage{graphicx}
\usepackage{subcaption}
\usepackage[outline]{contour}
\usepackage{comment}
\usepackage{floatrow}
\usepackage{float}
\definecolor{maskblue}{HTML}{1ebbee}

\title{Text and Click inputs for unambiguous open vocabulary instance segmentation}

\addauthor{Nikolai Warner\textsuperscript{*}}{nwarner30@gatech.edu}{1}
\addauthor{Meera Hahn}{meerahahn@google.com}{2}
\addauthor{Jonathan Huang}{jonathanhuang@google.com}{2}
\addauthor{Irfan Essa}{irfan@gatech.edu}{1,2}
\addauthor{Vighnesh Birodkar}{vighneshb@google.com}{2}

\addinstitution{
 Machine Learning Center\\
 Georgia Institute of Technology\\
 Atlanta, GA
}
\addinstitution{
 Google, Inc.\\
 Mountain View, CA
}
\runninghead{Warner et al.}{Text and Click for open vocabulary segmentation}

\begin{document}

\maketitle
{\footnotesize \textsuperscript{*} Work performed partially while at an internship at Google.}
\begin{abstract}

Segmentation localizes objects in an image on a fine-grained per-pixel scale. Segmentation benefits by humans-in-the-loop to provide additional input of objects to segment using a combination of foreground or background clicks. Tasks include photo-editing or novel dataset annotation, where human annotators leverage an existing segmentation model instead of drawing raw pixel level annotations. We propose a new segmentation process, Text + Click segmentation, where a model takes as input an image, a text phrase describing a class to segment, and a single foreground click specifying the instance to segment. Compared to previous approaches, we leverage open-vocabulary image-text models to support a wide-range
of text prompts.
Conditioning segmentations on text prompts improves the accuracy of segmentations on novel or unseen classes. We demonstrate that the combination of a single user-specified foreground click and a text prompt allows a model to better disambiguate overlapping or co-occurring semantic categories, such as ``tie'', ``suit'', and ``person''. We study these results across common segmentation datasets such as refCOCO, COCO, VOC, and OpenImages. Source code available \href{https://github.com/nikolaiwarner7/Text-and-Click-for-Open-Vocabulary-Segmentation}{here}.\vspace{-2mm}
\end{abstract}

-------------------------------------------------------------------------
\section{Introduction}
\label{sec:intro}

\vspace{-2mm}
Instance segmentation is
the problem of labelling every
single pixel that belongs
to a known set of categories.
Deep-learning based methods
 have shown tremendous progress
in recent years with early works
such as Mask R-CNN \cite{maskrcnn}
and more recently with 
Cascade R-CNN 
\cite{cascadercnn}, SOLOv2 \cite{solov2} and
MaskFormer \cite{maskformer}. 
Although broadly applicable when we have a lot
of labeled data, fully supervised instance
segmentation methods are limited to the 
set of categories they are trained on.
In this paper, we explore a model that can
be more useful by taking inputs from the user
about what objects to segment.
We ask for 2 inputs: (i) a single click on the object to be segmented and (ii) a text description of the same object.

\begin{figure}[t!]
\centering
\begin{subfigure}{0.22\textwidth}
\includegraphics[width=\linewidth]{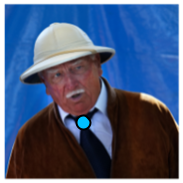}
\caption{Input and  click({\color{maskblue}\textbullet})}
\label{fig:click_amb_input}
\end{subfigure}
~
\begin{subfigure}{0.22\textwidth}
\includegraphics[width=\linewidth]{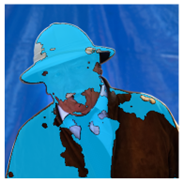}
\caption{No-text baseline}
\label{fig:click_amb_baseline_pred}
\end{subfigure}
~
\begin{subfigure}{0.22\textwidth}
\includegraphics[width=\linewidth]{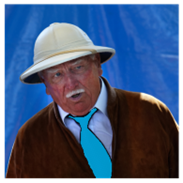}
\caption{Text $=$ ``Tie''}

\label{fig:click_tie_pred}
\end{subfigure}
~
\begin{subfigure}{0.22\textwidth}
\includegraphics[width=\linewidth]{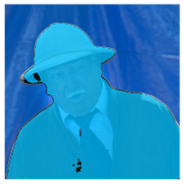}
\caption{Text $=$ ``Person''}
\label{fig:click_person_pred}
\end{subfigure}\vspace{-2mm}
\caption{The benefit of text input for instance
segmentation. 
The model in \ref{fig:click_amb_baseline_pred}
struggles to guess the correct object based
on only the point input from \ref{fig:click_amb_input}.
Our approach, which takes both text and click as input
is successfully able to segment \ref{fig:click_tie_pred} and \ref{fig:click_person_pred}. Both models
are trained on OpenImages with 64 seen classes.\vspace{-2mm}
}
\label{fig:click_amb}
\end{figure}

In isolation, each of these modalities is insufficient
to unambiguously designate a single instance to be 
segmented.
For
example, consider the click in Figure \ref{fig:click_amb_input}. It is unclear what the user wants to segment based on
this one input. The user could mean they
want to select the whole person or just
the tie or shirt. This lack of specificity
is also reflected in a model trained
on single-click data, as seen in Figure
\ref{fig:click_amb_baseline_pred}.
Similarly, text input alone can also be ambiguous
--- for example, using ``car'' as text input would be 
insufficient to describe a single instance if there
are multiple cars in an image.  Though there are ways
to address this ambiguity through the use of
referring expressions~\cite{kazemzadeh2014referitgame,mao2016generation,refcoco2016,referringExpression}, these approaches place a
heavy burden
on the user to carefully construct perfectly
unambiguous text phrases.  
Together however, a Click + Text input mechanism
is a simple low-effort way to unambiguously
designate an instance in an image to be segmented.

A similar framework was first delineated by the
PhraseClick~\cite{phraseclick} paper, which 
proposed an architecture that 
takes text as input using a bi-directional
LSTM.
Although  PhraseClick addresses the
ambiguity problem, it does so in a class
specific manner. Their approach only
learns to model the classes in their
training dataset, and has no way to
generalize beyond the set of words
that it sees during training. 

Our model uses the same set
of inputs as PhraseClick (Click + Text),
but goes  beyond the fixed set
of words it observes during training.
To do so, we leverage the generalization
abilities of image-text models
such as CLIP~\cite{clip},
which have demonstrated  zero-shot
generalization abilities by learning
from web-scale image/text pairs. 
Specifically, our method relies on
saliency maps 
extracted from CLIP style models (e.g. using recent approaches such as
MaskCLIP~\cite{MaskclipZhou} or Transformer Explainability by Chefer et al~\cite{chefer2021generic}).
These ``text saliency'' methods allow us to  gauge the relevance of 
each pixel in an image to a given text-query. 
Because models like CLIP~\cite{clip} are
trained on large, open-vocabulary datasets,
approaches like MaskCLIP~\cite{MaskclipZhou} gives
us a coarse, semantic-level understanding
of a wide variety of concepts (see heatmap examples in Appendix).
And combined with a click from the user, this gives us precise
information about which instance they want to segment.

The benefit of using text and click can be seen 
in Figure \ref{fig:click_tie_pred} and
\ref{fig:click_person_pred}.  Our model
can successfully
use the input text to predict 2 different
objects given the same input point, 
and it is able to do so for text inputs
 beyond the categories it has seen during
training time. 

Our main contributions are as follows:\vspace{-2mm}
\begin{enumerate}
  \setlength{\itemsep}{1pt}
  \setlength{\parskip}{0pt}
  \setlength{\parsep}{0pt}
    \item We propose to condition segmentation models on text by
    leveraging pre-trained CLIP models
    using MaskCLIP to
    generate a per-pixel saliency that is used as input to our model
    and show  our approach to be effective for novel category generalization.
    \item We show that our approach matches or exceeds
    the performance of the PhraseClick method~\cite{phraseclick} 
    while generalizing to many more categories.
    \item We compare with the recent Segment Anything (SAM)~\cite{segany} model and show that we outperform it on the task of segmenting
    instances based on single click and text
    as input, while training on a much smaller dataset.
\end{enumerate}

\begin{figure}[!ht]
\centering
\subfloat[Our prediction for 
`Model globe' and `Basket']{\includegraphics[width=.65\textwidth]{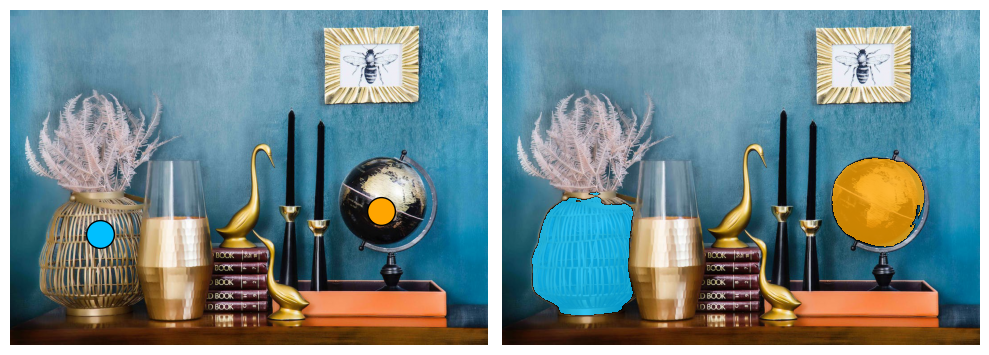}}\\
\subfloat[Our prediction for `Kayak paddle' and `Helmet']{\includegraphics[width=.65\textwidth]{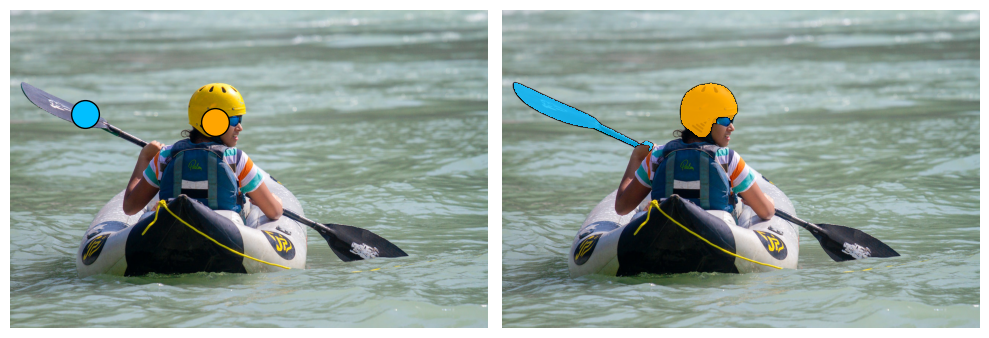}}\\
\subfloat[Our prediction for `Microscope' and `Hairnet']{\includegraphics[width=.65\textwidth]{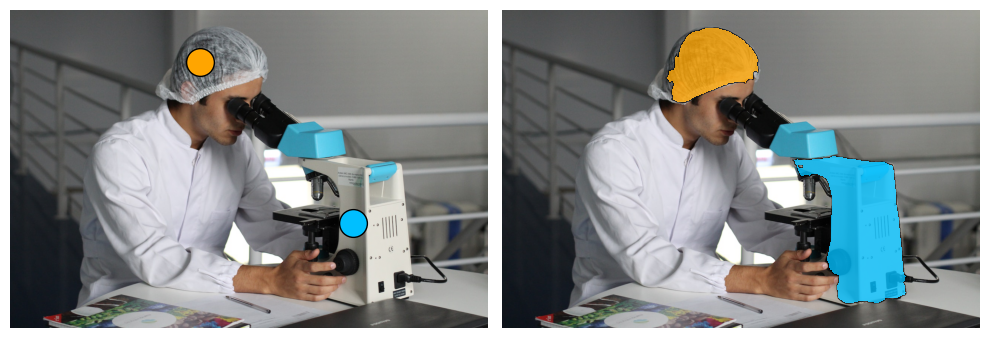}}\vspace{-1mm}
\caption{Open vocabulary queries demonstrated on images from the web. These categories include `Kayak Paddle', `Basket', and `Microscope' which are never seen by the model.}\vspace{-2mm}
\label{fig:Inference Web Images}
\end{figure}

We also experiment with truly open-vocabulary setting on queries far out of distribution from academic datasets. As evident in Figure \ref{fig:Inference Web Images}, our model performs well on classes that were outside of seen or unseen sets within the training data, on images completely distinct from our training or validation data.


\section{Related Work}
\label{sec:rel_work}

\noindent\textbf{Semantic / Instance Segmentation.} Semantic segmentation is the problem of assigning a semantic label to each pixel in an image~\cite{MO2022626}. Because it requires a large dataset of dense annotations however, it can be time-consuming and expensive to crowd-source. Training segmentation models in new or niche domains therefore is constrained by data annotation availability and cost. State of the art semantic segmentation techniques employ a fully convolutional architecture that combine low level and high level feature maps for accurate segmentation masks~\cite{HAO2020302}.  Deeplab V3~\cite{deeplab} uses atrous convolutions to capture objects and features at multiple scales spanning large and small and its successor  DeeplabV3+~\cite{chen2018encoderdecoder} remains a strong SOTA segmentation architecture,  adding a decoder module to Deeplab V3 to improve segmentation quality along object boundaries. Another class of state of the art segmentation models are  based on Vision Transformers (or ViT)~\cite{dosovitskiy2020image},
and extend it to segmentation by decoding  image patch embeddings from ViT to obtain class labels  (e.g., \cite{strudel2021segmenter}) This family includes
SegViT \cite{zhang2022segvit} that proposes to better use the attention mechanisms of ViT to generate mask proposals, as well as ViT-Adapter-L~\cite{chen2022vision} that attempts to correct weak priors in ViT using a pre-training-free adapter.

\noindent\textbf{Interactive Object Segmentation.} Interactive object segmentation seeks to utilize additionalhuman inputs such as clicks or bounding boxes at inference time to guide/refine a segmentation.
Deep interactive object detection~\cite{deep_int_object_xu} use a novel strategy to select foreground and background points from an image, which are transformed via Euclidean distance maps in to channels that can be 
used as inputs into a convolutional network. PhraseClick\cite{phraseclick} explores how to produce interactive segmentation masks using text phrases in a fully supervised manner as an additional modality of input. They demonstrate that adding phrase information reduces the number of interactions required to achieve a given segmentation performance, as measured by mIoU.

Sofiiuk et al.~\cite{reviving_iterative} highlights the issue with other inference-time optimization procedures in related works and proposes an iterative training procedure with a simple feedforward model. Focal click\cite{chen2022focalclick} highlights how existing interactive segmentation models can perform poorly on mask refinement when they destroy the correct parts; and proposes a new method that refines masks in localized areas. SimpleClick \cite{liu2022simpleclick} explores ViT in the context of interactive segmentation, adding only a patch embedding layer to encode user clicks without extensively modifying the ViT backbone. 

\noindent\textbf{Zero Shot Segmentation.}
ZS3Net\cite{zero_shot_sem_seg} 
performs zero shot semantic 
segmentation by correlating
visual and text features using word2vec~\cite{wordtovec}. They also introduce a self-training procedure using pseudo-labels for pixels of unseen classes.
CAGNet~\cite{CAGnet}  adds a contextual module that takes as input the segmentation backbone output and predicts a pixel-wise feature and contextual latent code per pixel. Their aim is to use more pixel-level information with their feature generator whereas ZS3Net contains a feature generator that uses only semantic word embeddings. \\
\indent While traditional end-to-end segmentation features are grouped implicitly in convolutional networks, GroupVIT~\cite{groupvit} seeks to explicitly semantically group  similar image regions into larger segments to perform zero-shot segmentation. It achieves 52.3\%  mIoU for zero shot accuracy on PASCAL VOC 2012. LSeg~\cite{LSeg2022} trains an image encoder to maximize similarity between the text embedding for a given query and the image embedding of the ground truth pixel classes. SPNet \cite{xian2019semantic} performs inference on unseen classes by utilizing semantic word embeddings trained on a free text corpus such as word2vec or fast-text. \\
\indent Zegformer \cite{DBLP:journals/corr/abs-2112-07910} achieves impressive results on zero-shot segmentation by ``decoupling'' the segmentation task into two stages: grouping pixels into likely segments in a class-agnostic manner, and assigning classes to grouped pixels. MaskCLIP~\cite{MaskclipZhou} achieved SOTA transductive zero-shot semantic segmentation by utilizing a pre-trained CLIP\cite{clip} model. They also showed that they can generate psuedo-labels of unseen categories and use it to train a semantic segmentation model. Although this approach can generalize to many classes, it necessitates training a new model for each set of new classes which is costly.

\section{Method}
\label{sec:method}

\begin{figure}[t]
  \includegraphics[width=.85\textwidth]{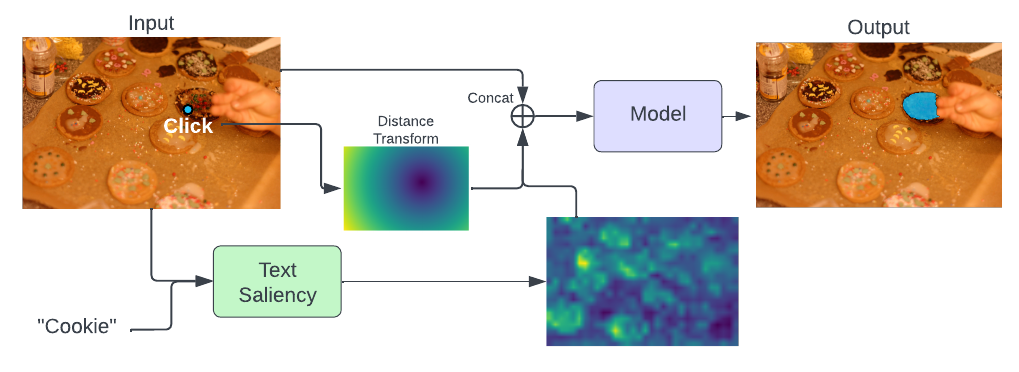}\vspace{-3mm}
  \caption[width=0.5\textwidth]{
  Our model architecture:
  we take as input a guided foreground click, the RGB image, and a text category. Then, the image-text saliency model (MaskCLIP here) produces a text-weighted feature map helping to localize the instance of interest. Finally, the original RGB image, clickmap, and saliency map are concatenated and fed into a modified fully convectional segmenation model, that accepts as input a 5 channel array.
  }
\label{fig:model_arch}
\end{figure}

Our main objective is to create a model capable of open vocabulary segmentation on novel classes. Figure~\ref{fig:model_arch} summarizes
our approach to this problem. We take as input to our segmentation model an RGB image, a single foreground click, and a text prompt and produces a class agnostic segmentation mask as output. While there are many possible ways to incorporate click and text cues into such a model, we take a simple but effective approach of encoding both side inputs as additional channels to be concatenated with the original input image, then fed to a standard segmentation network (e.g., DeepLabV3+, which we use in our experiments). Specifically, our foreground click is passed through a Euclidean distance transform to create a map with a continuous range of values normalized to [0, 1]. This is a standard  technique in the interactive segmentation literature~\cite{deep_int_object_xu}.

\begin{figure}
    \centering
    \subfloat[Input Image]{
        \includegraphics[width=0.2\textwidth]{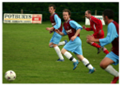}
    }
    \subfloat[``Ball'' heatmap]{
        \includegraphics[width=0.2\textwidth]{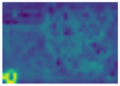}
    }
    \subfloat[``Shirt'' heatmap]{
        \includegraphics[width=0.2\textwidth]{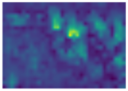}
    }\vspace{-2mm}
    \caption{Example
    of text saliency heatmaps
    produced by MaskCLIP~\cite{MaskclipZhou}. The
    heatmaps give us a rough estimate of where the
    input text is localized, while supporting
    the large vocabulary learnt by CLIP~\cite{clip}.
    }
    \label{fig:heatmap_demo}
\end{figure}

In order to convert a text prompt to a single channel image, we passed the text prompt
through a text-saliency model to produce a spatially sensitive guess (i.e., a saliency heatmap) of what pixels are similar to a given text query.  In our experiments, we use the MaskCLIP text-saliency model model~\cite{MaskclipZhou} which allows us to effectively incorporate a textually-sensitive, spatial saliency map that takes as input any open vocabulary text prompt. We note that MaskCLIP builds on the CLIP vision-language model that learns to align similar images and text queries via its massive web-scale dataset of image-caption pairs and contrastive learning scheme. In Figure~\ref{fig:heatmap_demo} we 
visualize the output of this method.

In our experiments, we have informally tried several saliency methods such as GradCAM 
~\cite{GradCAM}, Generic Transformer Interpretability ~\cite{chefer2021generic}, and  MaskCLIP~\cite{MaskclipZhou}. 
From qualitative experiments, we observed the best results from MaskCLIP,
and it also represents a strong baseline that is easy to implement with a few changes
to the encoder layer of CLIP.  

Our choice of converting a text prompt to a single channel image is nonstandard; however, we argue that it has a number of benefits. In using a powerful text-saliency model, we significantly lessen the burden on our own segmentation network since its  task can now be viewed as that of refining a (admittedly) rough initial segmentation into a clean segmentation given the image and click. Moreover, since this saliency
heatmap representation is itself class agnostic, our network should conceptually generalize well to classes that it did not get to see at training time (and we show that this is indeed the case in our experiments). As a contrast, the PhraseClick paper~\cite{phraseclick} embeds text inputs with Word2Vec and uses a bidirectional LSTM to model contextual relations between words in a phrase. However their image and text vector representations are not explicitly aligned; the image embedding vector is simply produced from a global pooling operation. Moreover, their model is not open-vocabulary, it is limited to a fixed set of prompts introduced during training.

\vspace{-3mm}

\section{Experiments}
\label{sec:experiment}

 


To measure our model's ability to generalize to novel classes, we train our model on a subset of all classes in the dataset (which we call ``seen classes''), but at test time evaluate the trained model on the remaining classes (called ``unseen classes'') as well as all classes present in the dataset. Where available (with the exception of OpenImages) we follow the standard zero-shot segmentation literature splits of ``seen'' and ``unseen'' classes in our experiments.  


In our experiments, we modify the first layer of a DeepLabv3+ model (with ResNet backbone) to accept a 5 channel image as input, and train all layers from scratch. We modify the number of output classes in the mask prediction module to 2 (to delineate foreground/ background) as we perform inference on each individual instance, and not all instances in a given image. We use  standard hyperparameters (based on the MMSeg implementation~\cite{mmseg2020})
for DeeplabV3+ 
and train on 2 Nvidia A40 GPUs with a batch size of 32. 
Our heatmaps and clickmaps are normalized per instance, to scale values between $[-1, 1]$.

To generate clicks for training, we sample a random point within the ground-truth segmentation mask boundary. Building off of
standard interactive segmentation 
literature~\cite{deep_int_object_xu}, positive points are selected to be at least some minimum distance from the object border, and a minimum distance from  other positive points. Negative points are sampled using a variety of strategies: first, from points near the border of the object mask boundary; second, from points in other object instances in the same image that we are not trying to segment.

We train separate models for the Pascal VOC~\cite{VOC2010}, COCO~\cite{COCO2014}, refCOCO~\cite{refcoco2016}, and OpenImages datasets~\cite{OIv4}. We train models in two configurations: zero-shot segmentation, and fully-supervised segmentation. In the former, the model has access only to instances in the limited set of seen-classes and RGB images that contain instances of those seen-class sets. For VOC, we use the 5 seen-class set defined in the ZS3~\cite{zero_shot_sem_seg} out of 20 total classes. For refCOCO and COCO, we use the standard 20/60 split of segmentation classes proposed in prior zero-shot segmentation literature. For our OpenImages experiments, we found no prior standard split for zero-shot segmentation, and there are 350 total segmentation classes. Thus, we use 
the intersection of the COCO classes and OpenImages segmentation classes as our seen set, resulting in 64 seen classes for training ($\sim 20\%$ 
of total classes). 
All results are reported at 90K iterations unless otherwise stated.

\subsection{Novel class generalization}
\label{sec:novel_class_gen}

In Table \ref{tab:ZSS all} we show that across all 4 datasets studied, conditioning on text-saliency improves overall mIoU across the board; and that this improvement mostly comes from larger improvements on the set of unseen classes. 
For example, on COCO, our heatmap-based model achieves 1.72 mIoU greater than  baseline on seen classes, but 6.98 mIoU greater than  baseline on unseen classes. In other words, the model is able to use the heatmaps to noticeably improve the quality of unseen class segmentations.

Moreover, the smaller the seen class set, 
the greater the benefit of conditioning the segmentation network on text saliency. We study this effect in Table~\ref{tab:OI Seen Class Ablation}, where we vary the fraction of classes designated as ``seen'' in the
OpenImages dataset. 
Here we see that the improvement increases as number of seen classes decreases; this intuitively makes sense as our technique of converting to a
saliency map places the main burden of novel class generalization
on the pretrained CLIP model rather than the segmentation network itself.


\begin{figure}[t!]
\center
\subfloat[RGB Input + Click]{\includegraphics[width=0.2\textwidth]{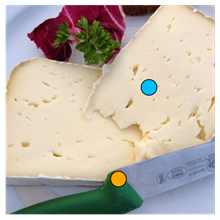}}
\subfloat[Baseline]{\includegraphics[width=0.2\textwidth]{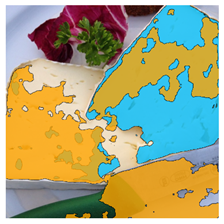}}
\subfloat[Ours]{\includegraphics[width=0.2\textwidth]{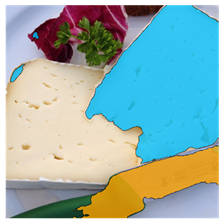}}
\vspace{0.5pt}
\subfloat[RGB Input + Click]{\includegraphics[width=0.2\textwidth]{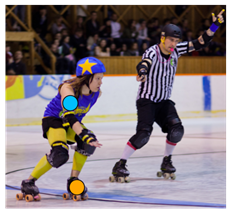}}
\subfloat[Baseline]{\includegraphics[width=0.2\textwidth]{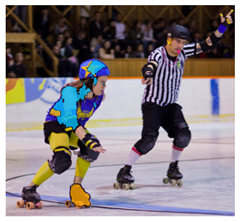}}
\subfloat[Ours]{\includegraphics[width=0.2\textwidth]{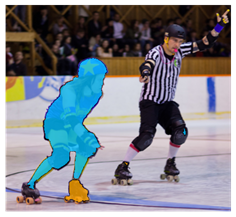}}
\vspace{0.5pt}
\subfloat[RGB Input + Click]{\hspace{-5pt}\includegraphics[width=0.20\textwidth]{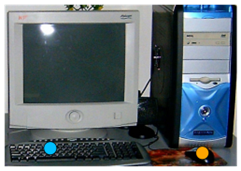}}
\subfloat[Baseline]{\includegraphics[width=0.20\textwidth]{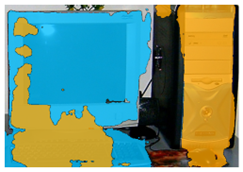}}
\subfloat[Ours]{\includegraphics[width=0.20\textwidth]{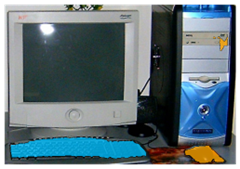}}
\caption{Inference examples on unseen classes for baseline versus our model for (a) ``Cheese'' and ``Knife'', (d) ``Roller Skates'' and ``Woman'', and (g) "Keyboard" and "Mouse". Conditioning on text saliency improves novel class segmentation and removes ambiguity. Model trained on OpenImages with 64 classes set as seen, compared to the click-only baseline without heatmaps.
}
\label{fig:Unseen Inference ZSS}
\end{figure}

\begin{table}[t!]
  \centering
  {\footnotesize
  \begin{tabular}{ lrrrrrr  }
  \toprule
    Dataset & Text Input & 
    \multicolumn{3}{c}{mIoU}\\
    \cmidrule{3-5}
    && Overall & Seen & Unseen \\

    \midrule
    refCOCO & \ding{51}  & 66.02 (+3.03) & 70.30 (+1.86) & 56.35 (+5.68) \\
    refCOCO & &  62.99 & 68.44 & 50.67 \\
    \midrule
    VOC & \ding{51}& 57.76 (+4.52) & 59.31 (+3.2) & 50.73 (+10.45) \\
    VOC &  &  53.24 & 56.11 & 40.28 \\
    \midrule
    COCO & \ding{51} & 38.42 (+3.89) & 42.06 (+1.72) & 33.45 (+6.98)   \\
    COCO & & 34.53 & 40.34 & 26.47  \\
    \midrule
    OpenImages & \ding{51}  & 57.05 (+4.40) & 67.03 (+3.35) & 53.92 (+4.74)  \\
    OpenImages &  & 52.65 & 63.68 & 49.18  \\
  \bottomrule
  \end{tabular}
  \caption{Results for Text+Click model
  on seen and unseen classes.
  We used one click for all models and
  trained using only seen classes. For
  OpenImages we use 64 seen classes. We convert text input  to a heatmap using Maskclip.}
  \label{tab:ZSS all}}
\end{table}
\begin{table}[t!]
  \centering
  {\footnotesize
  \begin{tabular}{ lrrrrl  }
  \toprule
    Seen Classes & Text Input  & 
    \multicolumn{3}{c}{mIoU}\\
    \cmidrule{3-5}
    && Overall & Seen & Unseen \\
    \midrule
    64 & \ding{51}& 57.05 (+4.4) & 67.03 (+3.35) & 53.92 (+4.74)\\
    64 &   & 52.65 & 63.68 & 49.18 \\
    \midrule
    34 & \ding{51} & 55.10 (+5.82) & 62.03 (+5.19) & 52.95 (+6.12)  \\
    34 & &49.28 & 56.84 & 46.83  \\
    \midrule
    23 & \ding{51}  & 53.65 (+7.89) & 61.64 (+8.38) & 51.14 (+7.62) \\
    23 &  & 45.86 & 53.26 & 43.53  \\
  \bottomrule
  \end{tabular}
  \caption{Difference in performance
  as the number of seen classes in OpenImages
  changes. Note that gap between our approach (Text+Click) and the click-only baseline
  increases with a smaller set of seen classes. We convert text input  to a heatmap using Maskclip.}
  \label{tab:OI Seen Class Ablation}
    }
\end{table}

\subsection{Qualitative examples}
In Figure \ref{fig:Unseen Inference ZSS} we provide several qualitative examples of our inference results.
In all of the examples, we click on unseen classes (e.g., ``cheese'', ``knife'', ``roller skates'', etc).
Here we use a model trained on OpenImages with 64 classes set as seen, 
and compare  to a simplified click-only baseline (same architecture) but without text saliency heatmaps
as input.
In the cheese and knife image for example, the baseline aims to separate object instances by features, 
but is confused by the overlapping textures from the cheese and knife instances. However, our model conditioned on text is able to clearly distinguish the separate cheese
instances and separate them from the knife.  



\subsection{Comparison with SAM}
\label{sec:compare_sam}

\begin{minipage}[t]{0.475\textwidth}
\footnotesize
{\footnotesize
  \centering
  \begin{tabular}[t]{ lrrrrr  }
  \toprule
        Dataset & \multicolumn{2}{c}{SAM}& 1-Click & Ours \\
        \cmidrule{2-3}
        & CLIP & Conf. \\
	\midrule
	COCO & 36.43 & 39.31 & 36.82 & \textbf{47.17} \\
  refCOCO & 47.07 & 52.48 & 66.16 & \textbf{68.07 }\\
  \bottomrule
  \end{tabular}
  \newline
  \newline
  }
\captionof{table}{
Comparing mIOU of our model with SAM\cite{segany}.
SAM outputs 3 predictions and we choose
one using SAM's confience (Conf.) or CLIP score(CLIP).
  Our models trained on all classes in COCO and
  refCOCO outperform SAM.}
\label{tab:sam_compare_all_classes} 

\end{minipage}
\hfill
\begin{minipage}[t]{0.475\textwidth}
\footnotesize
  \begin{tabular}[t]{ lrrrrr  }
  \toprule
    Dataset & Model & 
    \multicolumn{3}{c}{mIoU}\\
    \cmidrule{3-5}
    && Overall & Seen & Unseen \\

    \midrule
    COCO & Ours   & 38.42 & \textbf{42.06} & 33.45 \\
    COCO & SAM & \textbf{39.31} & 41.73 & \textbf{37.59}  \\
    \midrule
    refCOCO & Ours & \textbf{66.02} & \textbf{70.30} & \textbf{56.35} \\
    refCOCO & SAM & 52.48 & 61.18 & 48.64 \\
    \midrule
    OI & Ours & 57.05 & \textbf{63.68} & 53.92  \\
    OI & SAM &\textbf{ 63.88} & 63.60 & \textbf{64.47}  \\
  \bottomrule
  \end{tabular}
\newline
\newline
 \captionof{table}{
Comparison with SAM while
  our model only trains on a subset
  of classes. Note that we outperform
  SAM on refCOCO.
  We use SAM's confience score to
  rank proposals in this experiment
  because it showed better results
  in Table~\ref{tab:sam_compare_all_classes}.
  OI=OpenImages.
}

\label{tab:sam_compare_unseen} 
\end{minipage}
\ \\
\ \\
The Segment Anything Model (SAM)~\cite{segany} is
a model that was trained 
with 1.1 billion masks from the SA-1B dataset.
SAM can work 
with a combination of positive/negative clicks
and text prompts and showed impressive
segmentation results with user-input.
In Table~\ref{tab:sam_compare_all_classes} we compare with SAM while 
taking as input a single class name and a click.
In spite of our smaller capacity
and limited data, we out-perform SAM when
training on all examples from COCO, refCOCO and
OpenImages. Note that a perfect apples-to-apples comparison is difficult here 
since SA-1B masks are not class-annotated  so we are not able to separate seen from unseen masks, given the SAM mode  an
unfair advantage in some ways. 
SAM outputs 3 predictions which we rank by CLIP scores or SAM's confidence scores. For CLIP scores we used the \texttt{ViT-L/14@336px} model, which SAM used in open-vocabulary training.\footnote{The text based open-vocabulary model was not made publicly available.}

In Table~\ref{tab:sam_compare_unseen}
we compare our approach with SAM while
only training on a subset of classes.
Note that even when we further limit our
training set and evaluate our model
on a set of classes that our model is \emph{guaranteed} to have not seen, 
we still outperform SAM on refCOCO.
It is important to note that
because of SAM's compute requirement,
we could not re-train SAM and only
evaluated the pre-trained model trained
on SA-1B.

\begin{figure}[!t]
\centering

\subfloat[RGB Input + Click]{\includegraphics[width=0.25\textwidth]{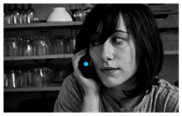}}
\subfloat[SAM]{\includegraphics[width=0.25\textwidth]{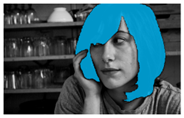}}
\subfloat[Ours]{\includegraphics[width=0.25\textwidth]{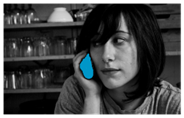}}
\vspace{1pt}
\subfloat[RGB Input + Click]{\includegraphics[width=0.25\textwidth]{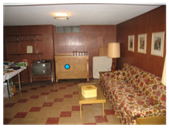}}
\subfloat[SAM]{\includegraphics[width=0.25\textwidth]{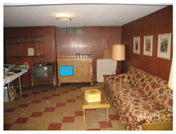}}
\subfloat[Ours]{\includegraphics[width=0.25\textwidth]{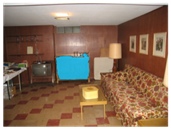}}\vspace{-2mm}
\caption{Example comparisons to SAM for (a) ``Mobile phone'', and (b) ``Chest of drawers''. Model trained on OpenImages with 64 classes set as seen, compared to SAM baseline using highest confidence prediction.
}\label{fig:SAM Comparisons}
\end{figure}

\section{Conclusion}
\label{sec:concl}
We set out to explore improved instance segmentation through the use of a single click and a text prompt. A single click is insufficient to specify what part of an instance to segment; a single text prompt can still be ambiguous unless carefully crafted. We have demonstrated that a single click combined with a text prompt outperforms a click-only baseline across a variety of datasets. We also show that a model conditioned on text-saliency can generalize much better to novel categories. We use  saliency maps from MaskCLIP to produce rough localizations for any  category. A separate segmentation model is trained on the concatenated input, and segments in a class-agnostic manner, while still retaining class-specific information from the MaskCLIP module. The recent SAM model is class-agnostic and struggles to disambiguate user intent on the overall part vs subpart from a single click.
Open vocabulary interactive segmentation is a novel task that has numerous applications, from reducing dense image annotation costs to improving background object removal in photo editing. We hope that the new text and click segmentation task will improve the accuracy of segmentations that require user interaction, while constraining the amount of interaction required. Future research directions could include automatically detecting the best category present around a user’s foreground click, to remove the necessity of an additional text input. Our work also intersects with research on how to produce refined segmentation masks from a rough or low quality input (bounding box, point, low quality mask). 

\bibliography{egbib}

\newpage
\section{Appendices}
\label{sec:Appdx}
\title{Text and Click inputs for unambiguous
open vocabulary instance segmentation}

\runninghead{Warner et al.}{Text and Click inputs for unambiguous open vocabulary instance segmentation}

\subsection{Details on data generation}
In our work, we compare interactive segmentation with a single click to segmentation conditioned on text saliency. The Phraseclick paper \cite{phraseclick} was the first paper to study combining a click and text query for disambiguation. In their experiments on refCOCO, they study different combinations of interactions, and the closest comparison to our work are the experiments they ran with 2 foreground clicks and a single background click. They cite previous work from Xu et al. \cite{deep_int_object_xu} for their experimental configuration. Negative background clicks are sampled either from other object instances present in a scene, from near the ground truth object boundary, or from anywhere that is not the ground truth object. Positive clicks are sampled subsequently with a minimum distance from each other. Finally, for every object instance in the ground truth dataset, random samples are taken to create augmentations in the training data  \cite{deep_int_object_xu}. The reccomended hyperparameters from Xu et al. include a 40 pixel minimum distance between sampled positive points, and 15 samples per each instance.

To focus on boundary quality of the generated interactive segmentations, we sampled negative points from two of the three strategies proposed by Xu et al. \cite{deep_int_object_xu}: from other instances of the same class present in a given scene, and from points along the outside boundary of the ground truth points. Additionally, we sampled positive points with 150 minumum distance from each other. We instead took a single sample per instance. This remains a future inquiry to see how the baseline model and ours conditioned on text saliency perform with additional data augmentation. From some anecdotal studies on refCOCO, the performance increase to both the baseline model and ours conditioned on text saliency is roughly 2-3 mIoU. For the fully supervised experiments in Table \ref{tab:Fully Supervised COCO, refCOCO}, we take 5 samples per instance for the much smaller dataset refCOCO, and a single sample per instance for COCO.

\subsection{Part Disambiguation and Visualizations for COCO}

\begin{figure}[!h]
\centering

\subfloat[RGB Input + Click]{\includegraphics[width=0.33\textwidth]{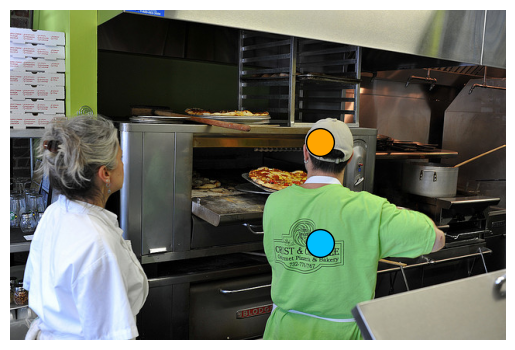}}
\subfloat[Baseline Prediction]{\includegraphics[width=0.33\textwidth]{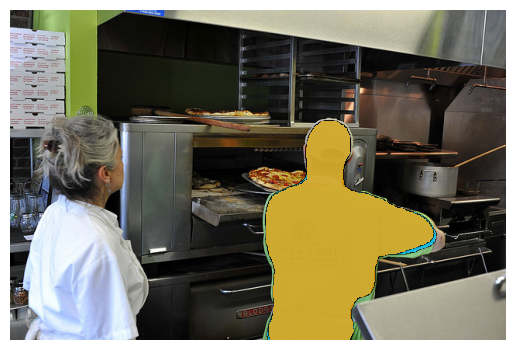}}
\subfloat[Our Prediction for `Shirt' and `Hat']{\includegraphics[width=0.33\textwidth]{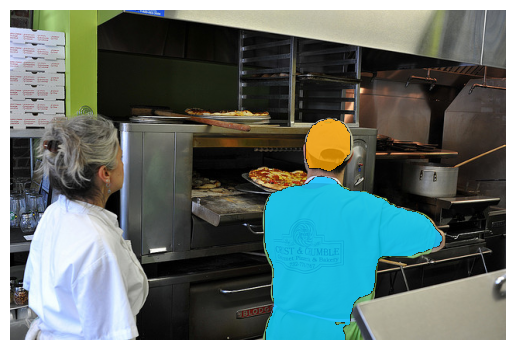}}
\newline
\subfloat[RGB Input + Click]{\includegraphics[width=0.33\textwidth]{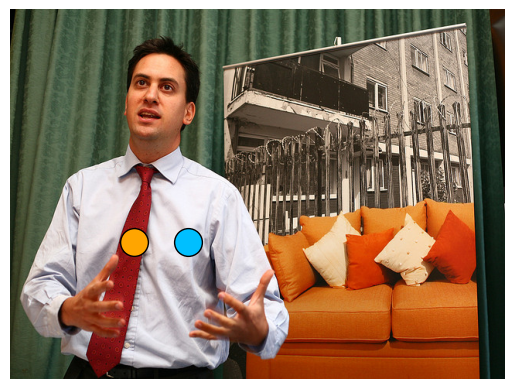}}
\subfloat[Baseline Prediction]{\includegraphics[width=0.33\textwidth]{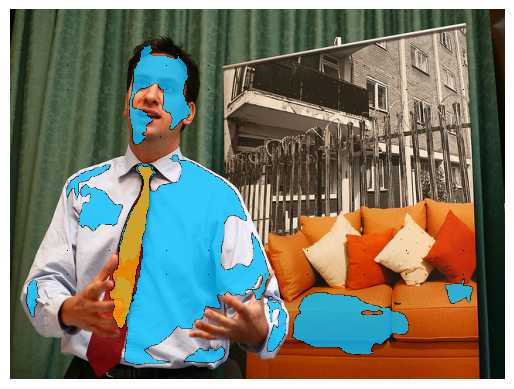}}
\subfloat[Our Prediction for `Tie' and `Person']{\includegraphics[width=0.33\textwidth]{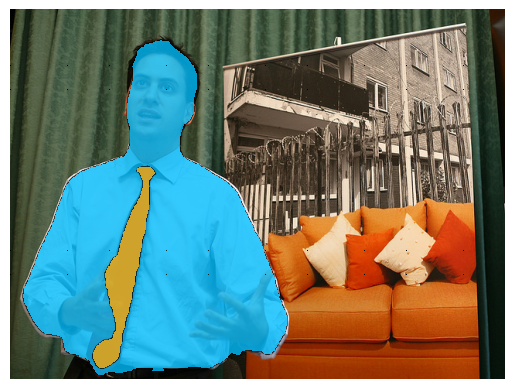}}
\caption{A comparison of the click-only baseline to text-saliency segmentation on the task of part disambiguation. The model here was trained in a zero-shot manner on OpenImages with 64 seen classes, and evaluated on validation images from COCO. The categories chosen are from the unseen class set. Text saliency conditioning helps the segmentation model disambiguate subparts such as the "tie" from the overall object of "person." Similarly, the segmentation model conditioned on text saliency is able to differentiate the classes of shirt, hat and person in the top row.
}\label{fig:ZSS Part Disambiguation Appendix}
\end{figure}

In our main paper, we demonstrate visual examples for the validation set of OpenImages~\cite{OIv4}. We also demonstrate in the Experiments section how text saliency improves novel class generalization. Here, we aim to provide examples of how the model trained on OpenImages performs on example images from COCO~\cite{COCO2014} in order to show model generalization across datasets.  As evident in Figure \ref{fig:ZSS Part Disambiguation Appendix}, our model qualitatively outperforms the baseline click only model in these settings as well. We show here that conditioning on text saliency also improves the ability of a model to generalize between a whole object and its sub-parts. This is also illustrated in the results shown in Figure \ref{fig:ZSS Part Disambiguation Appendix}.

\subsection{Experiments in a Fully Supervised Setting}

We present results of a fully-supervised version of our text-and-click model and these results are shown in Table \ref{tab:Fully Supervised COCO, refCOCO}. The purpose of Table \ref{tab:Fully Supervised COCO, refCOCO} is to understand the delta between the zero-shot segmentation model and a model trained in a fully-supervised manner. 
Whereas our zero shot segmentation model achieved 66.02 mIoU with a single click and text prompt on refCOCO, our fully supervised model in \ref{tab:Fully Supervised COCO, refCOCO} achieves 68.07 mIoU with a single foreground click and text prompt, and 72.89 with two foreground clicks, one background click and a text prompt. 

This demonstrates that constraining the model to a limited set of classes does not lead to a significant performance drops, we attribute this to our text saliency conditioning. The baseline zero shot segmentation refCOCO model with only a single foreground click achieved 62.99 mIoU, indicating a more notable performance drop versus the model conditioned on text saliency. Similar results are seen for COCO in Figure \ref{tab:Fully Supervised COCO, refCOCO} where the model trained in fully supervised mode achieved 47.17 mIoU for a single foreground click and text prompt; our zero shot model trained on only 20 seen classes achieves only 38.42. 

Table \ref{tab:Fully Supervised COCO, refCOCO} additionally contains an ablation experiment which ablates the number of interaction signals received by the model and determines how performance varies. We establish that inputting a text prompt counts as an interaction, and we therefore compare various configurations of text prompt inputs, foreground and background clicks. We hypothesized that additional interactions would decrease the utility of text saliency, because the object or subpart to segment would already be clearly defined. We find this generally confirmed, especially for results on the COCO dataset. We can see this when comparing rows 1-2 and rows 3-4 in Figure \ref{tab:Fully Supervised COCO, refCOCO}: under the condition of a single foreground click, the addition of a text prompt boosts mIoU by 10.35 for COCO; whereas under the condition of 2 foreground clicks and a single background click, the addition of a text prompt only boosts mIoU by 2.19. 

\begin{table}[t!]
  \centering
  \begin{tabular}{ lllrr }
  \toprule
    \multicolumn{3}{c}{Interaction}     &  \multicolumn{2}{c}{Overall mIoU}    \\
    \cmidrule{0-2} 
    Text Input & Pclicks & Nclicks  & COCO & refCOCO
    \\
    \midrule
     \ding{51} & 2 & 1 & \textbf{54.46} & \textbf{72.89} \\
    & 2 & 1 & 52.27 & 71.53 \\ 
    \ding{51} & 1 & 0 & \textbf{47.17}  & \textbf{68.07}  \\
     & 1 & 0 & 36.82 & 66.16 \\
    & 1 & 1  & 47.13 & 66.23  \\
  \bottomrule
  \end{tabular}
  \newline
  \newline
  \caption{Results of our model in the fully supervised setting over the COCO and refCOCO datasets for the text-click instance segmentation task. We convert text input  to a heatmap using Maskclip. The left hand side of the table shows the number of inputs given to the model in terms of text-saliency heatmaps, positive clicks (PClicks) and negative clicks (NClicks). The interaction setting with the highest mIoU is bolded for reference.} 
  \label{tab:Fully Supervised COCO, refCOCO}

\end{table}

\subsection{Comparison with SAM Model}
In our experiments, we compared our model to the Segment Anything Model (SAM) from Meta AI. Performing comparisons on the zero-shot segmentation setting was infeasible due to the extremely large number of GPUs required to retrain. To do so would require retraining the SAM model on a limited number of seen classes in its dataset. To the best of our knowledge, data category labels were not available at training time. In lieu of these experiments, we  conduct comparisons to the pre-trained SAM network. 

In Table \ref{tab:sam_compare_all_classes}, we explained that SAM outputs multiple mask proposals. We explore multiple strategies to filter their mask proposals to the best available proposal. We previously discussed using the CLIP similarity of each mask proposal crop to the ground truth text prompt. This was meant to produce an even comparison, since our model has access to the ground truth category label of an instance to generate the text saliency map. We also discussed the SAM confidence score. For sake of thoroughness, we also re-implement the Oracle score described in the SAM paper. They note that their model can be penalized by automated evaluation metrics because it suggests multiple masks; and note that the model produces SOTA results if allowed to compare its mask proposals to the ground truth one. See the Experiment sectionin the main paper for details. This suggests that the SAM model struggles with disambiguating mask proposals, though it can often suggest high quality ones. 

\subsection{Looking at Distractors and Neighboring Object Segmentation}

We analyze the role that distracting objects play in generating interactive segmentation. A given image can have multiple classes present, for example a table and a lamp. For a given class, multiple instances can be present, for example, a parking lot with multiple instances of the class `vehicle'.
Since the model learns to segment guided by a click and an open-vocabulary salience map (generated from a text category). This becomes a more challenging task the more objects and instances that are present, particularly the closer they are in proximity. We achieve the best results on refCOCO, a subset of COCO data re-sampled to make human annotation easier.

\begin{figure}[t]
  \includegraphics[width=.75\textwidth]{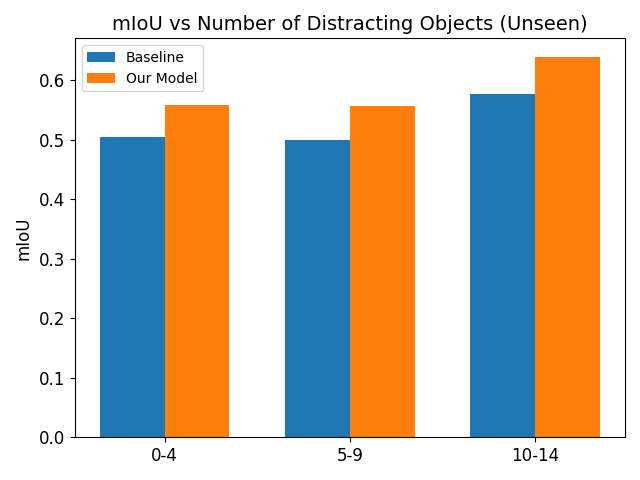}
  \caption[width=0.5\textwidth]{Bar chart displaying the effect of number of objects on segmentation quality. Using the OpenImages dataset we plot the average mIoU for images with N number of objects of the same class present. This analysis is only for instances of classes not seen during training. The model is trained on OpenImages with 64 classes as seen, and evaluated on the OpenImages validation set with all classes available. The baseline here is the click-only model compare to ours conditioned on text saliency. Our model consistently outperforms the baseline model, though the gap is similar across the number of distracting objcets present.}
\label{fig:Distractor Study}
\end{figure}

Results area available in Figure \ref{fig:Distractor Study}. In this experiment, we measure the number of instances of the same class present in a given image, and record the mIoU for each instance in the validation set along with the number of distractors present, only for instances of unseen classes. Our model consistently outperforms the baseline model, though the gap is similar across the number of distracting objcets present.

\subsection{Limitations and Future Work}

\begin{figure}[!ht]
\centering

\subfloat[RGB Input + Click]{\includegraphics[width=0.25\textwidth]{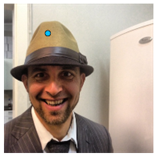}}
\subfloat[Heatmap]{\includegraphics[width=0.25\textwidth]{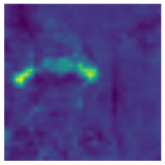}}
\subfloat[Baseline Seg]{\includegraphics[width=0.25\textwidth]{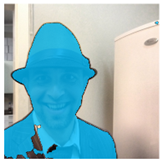}}
\subfloat[Our Seg]{\includegraphics[width=0.25\textwidth]{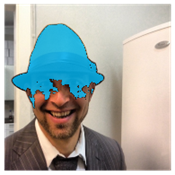}}
\newline
\subfloat[RGB Input + Click]{\includegraphics[width=0.25\textwidth]{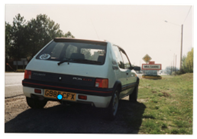}}
\subfloat[Heatmap]{\includegraphics[width=0.25\textwidth]{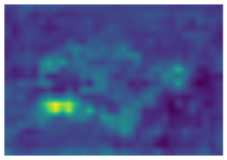}}
\subfloat[Baseline Seg]{\includegraphics[width=0.25\textwidth]{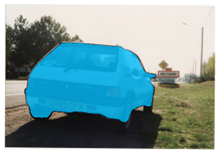}}
\subfloat[Our Seg]{\includegraphics[width=0.25\textwidth]{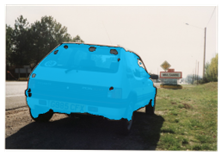}}
\newline
\subfloat[RGB Input + Click]{\includegraphics[width=0.25\textwidth]{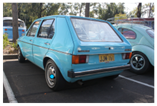}}
\subfloat[Heatmap]{\includegraphics[width=0.25\textwidth]{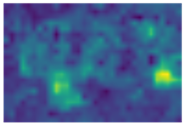}}
\subfloat[Baseline Seg]{\includegraphics[width=0.25\textwidth]{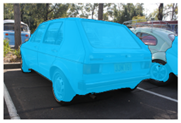}}
\subfloat[Our Seg]{\includegraphics[width=0.25\textwidth]{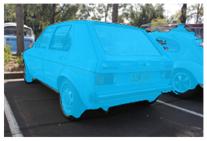}}
\newline
\newline
\caption{Examples of failure cases of text-click instance segmentation of our model. These examples show instances where despite the text saliency localizing the object of interest, the segmentation mask fails. We observe this largely happens in overlapping objects or when the queried category is a sub-part of a larger object. Categories: (a) "Cowboy hat", (b) "Vehicle registration plate", (c) "Wheel".
\newline
}\label{fig:Failure examples}
\end{figure}

We identify two main failure modes of the proposed model. The first instance results as a cascading error in cases with a low quality heatmap. In our experiments, we tried a few saliency techniques including GradCAM \cite{GradCAM}, Chefer 2021 et al. \cite{chefer2021generic} and MaskCLIP \cite{MaskclipZhou}. We found MaskCLIP to qualitatively perform the best, but improved saliency maps remains an important future line of inquiry. Sometimes, the heatmap helps to localize a given text query, but the segmentation network we train still fails to accurately segment it. We can see this second failure mode illustrated in Figure \ref{fig:Failure examples}. In the example in the bottom row of the figure, the heatmap has reasonably high probability over the pixels of the car wheel however the predicted segmentation contains the pixels of the entire car. Similarly, in the example in top row - containing an instance segmentation for a hat - the heatmap for the hat is high quality, but it predicts part of his whole body. We suspect that this is due to an imbalance of annotations in the training data; there are plenty of instances of whole objects such as automobiles or entire person silhouettes, but very few of a wheel, license plate, or hat.

\subsection{Comparison to Interactive Click Methods}


\begin{table}[h]
\footnotesize
\begin{center}
\resizebox{\columnwidth}{!}{  
  \begin{tabular}{ lrrrrrr }
  \toprule
    Model & Train Data & Eval Data & 
    \multicolumn{3}{c}{mIoU}\\
    \cmidrule{4-6}
    &&& Overall & Seen & Unseen \\
    \midrule
    RITM~\cite{sofiiuk2022reviving} & SBD & COCO Validation   & 38.86 & 45.00 & 30.33\\
    Ours & COCO[voc classes] & COCO Validation & 38.42  & 42.06 & 33.45  \\
    \midrule
    RITM~\cite{sofiiuk2022reviving} & SBD & OpenImages  & 49.42 & 67.39 & 47.17 \\
    Ours & COCO[voc classes] & OpenImages & 44.55 & 53.16 & 41.87  \\
  \bottomrule
  \end{tabular}
}
\caption{Comparing generalization of models to unseen classes. Comparison of RITM interactive segmentation model with a single click and our model with a click + text label. Using RITM checkpoint trained on all VOC classes with the SBD data. Our model was trained on VOC classes of COCO. COCO validation has 80 classes, and OpenImages has 300 classes.
We will include a comparison while training on SBD in the camera ready.}
\label{tab:ritm_compare_unseen} 
\end{center}
\end{table}

In Segment Anything~\cite{segany} Sec 7.1, Krillov et al. compare SAM to other interactive segmentation baselines (RITM, SimpleClick, and FocalClick) on single-click segmentation across 23 datasets. In Figure 9c and 9d  of\cite{segany}, SAM significantly outperforms all other methods on a single click, though the gap is much smaller for 2,3, or 5 clicks. This is because many interactive segmentation models are trained for mask refinement as opposed to generating the optimal proposal from a single click. These papers report the number of clicks to achieve a target IoU, but we are interested in minimizing interactions by combining text and clicks. We compare to RITM~\cite{sofiiuk2022reviving} in Tab.~\ref{tab:ritm_compare_unseen}, and show better generalization on COCO when trained on 20 VOC classes to unseen COCO classes. RITM is stronger on OpenImages, but the gap between seen and unseen is larger, suggesting RITM is a stronger click baseline but generalization could benefit from text saliency conditioning.

\subsection{Boundary IoU Metrics}


Please see table \ref{tab:boundary}. As you can see, using boundary IoU instead of mIoU, we achieve similar results, where our model slightly beats SAM on COCO validation.z

\begin{table}[h]
\footnotesize
\begin{center}
\resizebox{.8\columnwidth}{!}{ 
\footnotesize
  \begin{tabular}[h]{ lrrrrr  }
  \toprule
    Dataset & Model & 
    \multicolumn{3}{c}{mIoU}\\
    \cmidrule{3-5}
    && Overall & Seen & Unseen \\
    \midrule
    COCO & Ours   & 39.62 & 40.51 & 38.47\\
    COCO & SAM~\cite{segany} & 38.93 & 37.63 & 40.65  \\
  \bottomrule
  \end{tabular}
}
\newline
\newline
\vspace{-5mm}
 \captionof{table}{
Boundary IoU comparisons on MS COCO.
}
\label{tab:boundary} 
\end{center}
\end{table}
\vspace{-7mm}

\vspace{5mm}

\subsection{Comparison to Referring Expression Segmentation Methods}

Our method  is able to generalize to unseen classes with text input
by using pre-trained CLIP. We show this
in the paper with our ``unseen'' metrics. PhraseClick, VLT and LAVT have 
no mechanism to do this and do no evaluate on unseen classes.

\begin{table}[h]
\footnotesize
\begin{center}
  \begin{tabular}{ lrr }
  \toprule
    Model & Input & mIOU    \\
    \midrule
    LAVT & Phrase & 72.73 \\
    VLT  & Phrase & 65.65 \\
    Ours & Class Name + 3 clicks & 72.89 \\
    Ours & Class Name + 1 click & 68.07 \\
  \bottomrule
  \end{tabular}
\caption{Comparison of our method to Referring expression segmentation algorithms. 
For 3 clicks we sample 2 foreground and 1 background click. For 1 click we sample 1 foreground click.
All models were trained on RefCOCO.}
\label{tab:refex_comp} 
\end{center}
\end{table}

The largest difference between our model, LAVT and VLT is that we can segment completely unseen
classes at test time.
We achieve 33.45 mIOU on 60 COCO unseen classes while training on
only 20 seen classes.
We unlock this capability by training on saliency maps from Maskclip that is able to leverage
all of the knowledge of a pre-trained CLIP model.
We compare with LAVT and LVT in fully supervised setting (all classes are seen) in Table \ref{tab:refex_comp} and show
that we are able to match their performance with 3 clicks.
LAVT and VLT have not published numbers from unseen classes.

Also LAVT and VLT ~\cite{VLT_Ding_2023} require more specific language than our model (``guy in black sitting  to left leaned over'' (Fig 6 ~\cite{yang2022lavt}) vs. ours - ``person''). We achieve similar performance with less specific text supervision. This is important since annotating referring expression datasets at scale is expensive. In contrast, our method only requires the much more readily available, ground-truth class.

\subsection{Comparison to Phraseclick}
The PhraseClick~\cite{phraseclick} was published before Vision-Language joint pretraining became a common method. Therefore, \cite{phraseclick} propose an attention attribute mechanism, whereby the visual features are global average pooled into the same common dimension as the embedding dimension for the text representation. Text input is processed using word2vec and a trainable bi-LSTM. 
The text input is not initially aligned with the distribution of visual features.
At inference time, if a novel query is presented, the PhraseClick model will be unable to use the text information to make an improved segmentation. 

In our work meanwhile, we use the Maskclip technique to produce a spatial saliency map, for any possible novel text query, that provides a rough guess for the location of a given query. Maskclip retains the explicit spatial information providing a useful initial guess to the location of an object. Our model is trained in a class agnostic manner after extracting a heatmap guess, and so learns to segment any given prompt. 


PhraseClick~\cite{phraseclick} did not release code nor model weights, we cannot provide visual comparisons.

\end{document}